\documentclass[letterpaper, 10 pt, conference]{ieeeconf}

\IEEEoverridecommandlockouts
\usepackage{algorithm}
\usepackage{algpseudocode}
\usepackage{amsmath, amssymb, amsfonts}
\usepackage{graphicx}
\usepackage{booktabs}
\usepackage{cite}
\usepackage{placeins}
\usepackage{url}

\newcommand{\Renderer}{\mathcal{R}}
\newcommand{\Mobs}{\mathcal{M}_{\mathrm{obs}}}
\newcommand{\Ainit}{\mathcal{A}_{0}}
\newcommand{\Astar}{\mathcal{A}^{*}}
\newcommand{\Mscaf}{\mathcal{M}_{\mathrm{scaf}}}
\newcommand{\Vcand}{\mathcal{V}_{0}}
\newcommand{\Vsel}{\mathcal{V}}
\newcommand{\Degov}{\mathcal{D}_{\mathrm{ego}}}
\newcommand{\method}{R2S-EGO}

\title{\LARGE \bf
R2S-EGO: Dual-Proxy Refinement for Sparse-Capture Real-to-Sim
}

\author{Shuai Fang$^{1}$, Xin Deng$^{1,2}$, Yuchen Kang$^{1}$,
Zhenjiang Li$^{1}$, and Jie Chen$^{1,\dagger}$\\
$^{1}$\textit{XPENG Robotics},
$^{2}$\textit{The Hong Kong Polytechnic University}%
\thanks{$^{\dagger}$Corresponding author: Jie Chen, \texttt{chenj81@xiaopeng.com}.}%
}

\makeatletter
\newcommand{\rIISittingTeaser}{%
  \begin{minipage}{\textwidth}
    \centering
    \includegraphics[page=5, trim=67 149 67 161, clip,
                     width=\textwidth]{material/figure_v2.pdf}
    \def\@captype{figure}%
    \caption{\textbf{Behavior-aligned third- and first-person sitting rollouts
    in simulation and on the real Unitree G1 humanoid.}
    Teal panels show the policy rollout in the R2S-EGO simulation scene;
    navy panels show the corresponding real-robot rollout, aligned at
    approach, reorientation, and sitting. The center columns compare the
    simulated ego view with the onboard head-camera view at each phase,
    showing closely matched appearance and behavior across the transfer.}
    \label{fig:sitting_teaser}
  \end{minipage}%
}
\makeatother

\IEEEaftertitletext{%
  \rIISittingTeaser
  \vspace{0.5\baselineskip}%
}

\begin{document}
\maketitle
\thispagestyle{empty}
\pagestyle{empty}


\begin{abstract}
Real-to-sim (R2S) depends on scene representations that render observations
along robot ego trajectories, yet dense multi-view capture limits per-environment
real-image capture-count efficiency, and sparse human capture can leave behavior-scoped robot views
under-supported. Camera-controlled synthesis can fill missing views, but its use
in R2S requires behavior-admissible queries and capture-anchored structural
conditioning. We present R2S-EGO, which couples a simulator-derived robot proxy
that represents the behavior-scoped executable query domain with a
capture-anchored geometry proxy that supplies scene-specific structural
conditions. Within this domain, fixed-budget selection targets current support
deficits for which geometry support is available. The generated observations
are assimilated as pseudo-observations to refine the visual asset, while real
captures remain anchors. The fused geometry proxy also supplies the scene
collision surface, which is refreshed between rounds. Together, these updates
refine the existing simulation scene while its robot dynamics and control stack
stay fixed. Across 48
frozen Unitree G1 ego views in three Replica scenes, six-view R2S-EGO reaches
19.062 dB PSNR,
compared with 14.226 dB for the strongest reported R2S baseline. Across five
paired policy-training seeds, R2S-EGO achieves $82.5\%\pm6.8\%$ real-G1 sitting
success, compared with $10.0\%\pm10.5\%$ for GaussGym.
\end{abstract}


\section{Introduction}

Real-to-sim (R2S) turns real environments into repeatable simulation settings
for robot training and evaluation
\cite{Xie2025Vid2Sim,Fu2025GaussGym,Li2024RoboGSim}. Its scenes are
commonly reconstructed from calibrated multi-view observations using neural or
Gaussian representations
\cite{Mildenhall2020NeRF,Kerbl2023,Qureshi2024SplatSim,Li2024RoboGSim}.
We focus specifically on per-environment real-image capture-count efficiency.
The existing simulator's visual asset is initialized from registered sparse
RGB captures. Human-convenient capture and behavior-scoped robot cameras need not
sample the same poses, visible surfaces, or occlusions; consequently, sparse
capture can leave behavior-scoped robot views with insufficient visual and
structural support. We refer to this failure mode as the
\emph{capture--consumption support gap}.

Camera-controlled novel-view and video synthesis can generate observations
beyond the captured views
\cite{Zheng2024CamI2V,He2024CameraCtrl,Wang2024MotionCtrl,Yu2024ViewCrafter}.
For R2S scene refinement, however, a synthetic observation is useful only when
it satisfies two distinct forms of validity. On the query side, the requested
camera sequence must be realizable by the embodied robot within the declared
behavior scope. On the evidence side, generation must be conditioned on
structure tied to the captured scene, because camera control alone does not
determine the layout, visibility, or occlusion of sparsely observed regions.
The central challenge is therefore to address the capture--consumption support
gap with observations whose queries are behavior-admissible and whose content
is conditioned by capture-anchored geometry. With limited generation resources,
a secondary prioritization step is required: the implementation ranks
under-supported queries for which the geometry proxy supplies usable structural
conditions.

R2S-EGO operationalizes behavior-scoped support with a simulator-derived
\emph{robot proxy}. Starting from a declared behavior scope, it rolls out a
fixed controller in simulation in every refinement round and maps each robot
configuration through the camera mount into a stream of executable ego-camera
poses. These poses represent the behavior-scoped query domain that the scene
must support. Initial conditions or reference motions are resampled across
rounds, and failed rollouts are discarded and resampled. Within this domain,
the current visual-asset deficit and geometry-proxy availability determine
which candidates are eligible, and temporal local maxima are retained as a
fixed budget of complementary keyframes. The robot proxy therefore identifies
which missing observations are behavior-relevant and physically realizable;
budgeted selection determines which eligible support deficits are addressed
first.

R2S-EGO addresses the structural requirement with a
\emph{geometry proxy}. This requirement distinguishes R2S scene refinement from
generic novel-view synthesis: a generated image can be perceptually plausible
yet place a surface or occluder inconsistently with the captured environment,
which matters when that image becomes a training observation at a calibrated
simulator pose. Registered captures first provide triangulated geometry
fragments and their multi-view support. A shape prior proposes structure that
is absent from those fragments, but initially lives in an independent
coordinate frame. R2S-EGO aligns each prior region in scale, rotation, and
translation using capture support, then fuses the aligned prior with observed
fragments and current asset geometry. The aligned prior is retained in its
object region, while NKSR~\cite{Huang2023NKSR} reconstructs regions without an
object prior.
Rendering this proxy along the camera path that ends at a selected
robot view produces frame-aligned cues for layout, visibility, depth ordering,
and occlusion. The
prior thereby extends structural context into sparsely observed regions while
registered capture anchors its scene-specific placement; the resulting
condition is a capture-aligned structural hypothesis for generation.

\begin{figure*}[!t]
  \centering
  \includegraphics[page=4, trim=80 188 91 184, clip,
                   width=\textwidth]{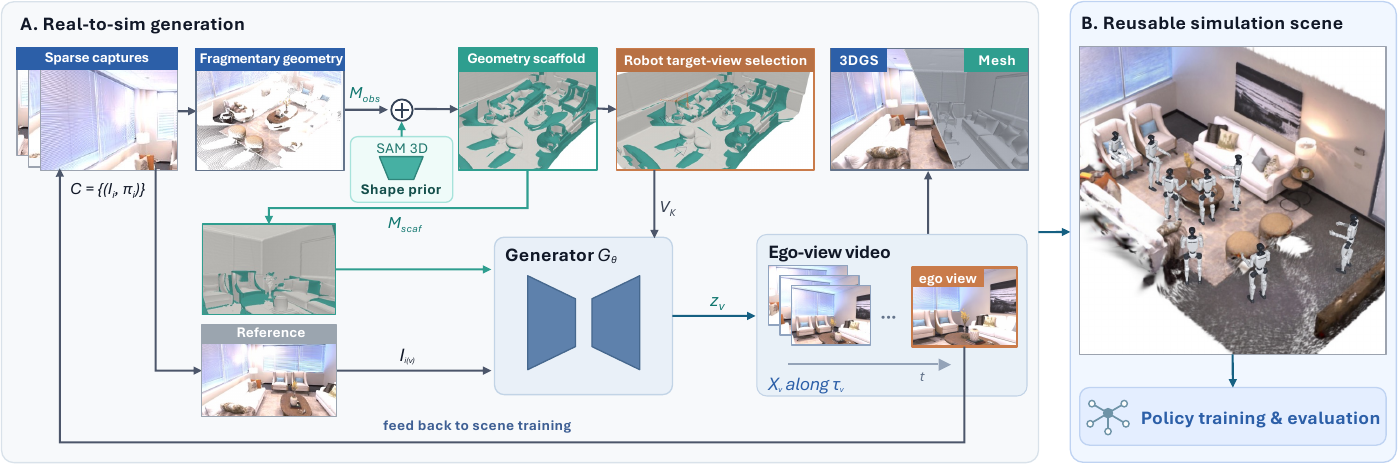}
  \caption{\textbf{R2S-EGO pipeline.}
  R2S-EGO refines an existing simulation scene along behavior-scoped robot
  views. A simulator-derived robot proxy represents executable camera queries
  and identifies which remain under-supported, while a capture-anchored
  geometry proxy supplies view-aligned structural conditions.
  Registered ego observations update the visual asset as lower-weight
  pseudo-observations while real captures remain anchors; the refreshed
  geometry/collision proxy supports the next-round rollout and downstream
  simulation.
  All panels show the same \texttt{room\_0} scene reconstructed from the same six
  captures. In the geometry panels, gray marks capture-observed surfaces and teal
  marks surfaces supplied by the shape prior.}
  \label{fig:pipeline}
\end{figure*}

The two proxies solve the R2S problem through their coupling
(Fig.~\ref{fig:pipeline}). During allocation, geometry availability prevents
the robot proxy from spending its budget on candidate poses that lack usable
structural support. During generation, the geometry proxy is rendered at the
same mounted-camera poses supplied by the robot proxy. A real reference image
provides scene appearance, relative camera motion specifies how the view
evolves, and the rendered proxy specifies what structure should be visible at
each pose. The generator consequently returns ego observations with estimated
camera poses: VGGT registers each retained frame relative to a fixed real
reference capture. R2S-EGO assimilates these frames into the visual
asset as lower-weight pseudo-observations while retaining registered real
captures as persistent anchors. This converts targeted generation into a
R2S scene-refinement loop: the observations update the visual asset,
while the fused geometry supplies the refreshed collision surface used by the
next-round rollout and downstream simulation.

We evaluate the resulting scene on held-out robot views and in downstream
policy use. Across 48 frozen Unitree G1 ego views in three Replica scenes, six-view
R2S-EGO reaches 19.062 dB PSNR, compared with 14.226 dB for the strongest
baseline. Under the controlled real-G1 sitting protocol, policies trained in
GaussGym and R2S-EGO simulation scenes succeed in 4/40 and 33/40 trials across
five independently trained policies per condition, respectively;
Fig.~\ref{fig:sitting_teaser} shows the R2S-EGO transfer, with
the simulated and onboard ego views compared at each sitting phase. Our
contributions are:
\begin{itemize}
  \item We formulate sparse-capture R2S scene refinement around the
        capture--consumption support gap, targeting support deficits within
        behavior-scoped executable robot views in an existing simulator rather
        than uniform global view completion.
  \item We introduce a dual-proxy method in which a robot proxy represents
        behavior-scoped executable camera queries and localizes current support
        deficits, while a capture-anchored geometry proxy supplies structural
        conditioning. Fixed-budget selection targets eligible deficits; their
        coupling then produces registered ego observations and a refreshed
        geometry/collision proxy for anchor-preserving R2S scene refinement.
  \item We validate the refined scene through frozen robot-view appearance
        evaluation and a controlled real-G1 downstream-policy comparison.
\end{itemize}


\section{Related Work}
\label{sec:related}

\paragraph*{Reconstructed and generated environments for robot learning}
R2S systems convert real observations into task-ready simulation assets.
RialTo and URDFormer recover interactive or articulated simulation assets
\cite{Torne2024RialTo,Chen2024URDFormer}, while SplatSim, RoboGSim, GaussGym,
Vid2Sim, Wanderland, and \mbox{RobotArena$^{\infty}$} connect reconstructed or video-derived
environments to robot training, transfer, or evaluation
\cite{Qureshi2024SplatSim,Li2024RoboGSim,Fu2025GaussGym,Xie2025Vid2Sim,
Liu2026Wanderland,Jangir2026RobotArena}.
Complementary generative models predict action-conditioned observations
\cite{Yang2024UniSim,Zhu2024IRASim}, or synthesize views for visual-policy
learning \cite{Yu2025LucidSim,Tian2025VISTA}. GRAIL likewise pairs 3D content
with video priors to obtain humanoid loco-manipulation on a G1
\cite{Xie2026GRAIL}, but its content is generated rather than
captured---text-to-mesh objects on procedural terrain---and its video prior
supplies motion, in that human--object interaction videos are converted into
body and object trajectories. R2S-EGO addresses a different
stage: it introduces neither a new simulator nor a policy learner, but performs
R2S scene refinement for a specific sparse-capture environment along
behavior-scoped robot views.

\paragraph*{Sparse scene recovery and generative refinement}
Sparse scene recovery spans radiance-field and Gaussian representations
\cite{Mildenhall2020NeRF,Kerbl2023}, few-view prior methods
\cite{Yu2021PixelNeRF,Chen2024MVSplat}, and pointmap-based geometric
reconstruction \cite{Wang2024DUSt3R,Leroy2024MASt3R}. 3DFIRES, WorldMirror,
and SAM~3D further infer structure beyond directly fitted visible surfaces
\cite{Jin2024ThreeDFIRES,Liu2025WorldMirror,Meta2025SAM3D}. In parallel,
video diffusion supplies appearance and temporal priors
\cite{Blattmann2023SVD,Xing2024DynamiCrafter}, while camera-controlled
generators synthesize calibrated novel-view sequences
\cite{Wang2024MotionCtrl,He2024CameraCtrl,Zheng2024CamI2V,Yu2024ViewCrafter},
and generated or restored views can be returned to NeRF or 3DGS optimization
\cite{Liu2024PseudoObservations,Kong2025GSGS,Zhong2025SceneGrounding,
Paliwal2025RI3D,Kim2025ExploreGS,Wu2025GenFusion,Xu2025FewGlimpses,
Nguyen2026PointmapDiff}. These methods are direct precedents for our
\emph{generated-view visual-update branch}: they improve novel-view rendering
by generating, repairing, or allocating pseudo views and feeding them back to
a radiance-field or Gaussian asset. Their stated task and evaluation, however,
end at reconstruction or visual camera exploration; they do not define or
evaluate an end-to-end robot simulation interface that couples the visual asset
to collision geometry, dynamics, embodied control, and downstream policy use.
Real-to-sim systems draw the same boundary: Vid2Sim combines Gaussian appearance
with mesh-based physical interaction, GaussGym synchronizes a 3DGS renderer with
collision physics, and Wanderland couples view synthesis with metric geometry
and collision meshes. R2S-EGO therefore does not claim novelty for feeding
generated views back into 3DGS. Its full output is instead a refined simulation
scene comprising an updated visual asset and refreshed geometry/collision proxy.
Its distinction is how generated views are selected and consumed: queries come
from a fixed robot embodiment, mounted camera, and declared behavior scope,
while capture-anchored structure constrains the views that become simulator
observations. This changes generic reconstruction coverage into targeted
support for behavior-scoped executable robot views used in downstream policy
training.

\paragraph*{Active and robot-targeted view allocation}
Next-best-view and active-mapping methods choose new sensing poses to improve
reconstruction or mapping
\cite{Pito1999NBV,Stachniss2005ActiveSLAM,Pan2022ActiveNeRF}. Recent systems
extend this principle with Gaussian-rendering information gain, feed-forward
uncertainty, vision-language guidance, or imagined geometry
\cite{Chen2025ActiveGAMER,Xu2026AREA3D,Li2026MAGICIAN}. These methods acquire
new real measurements for global reconstruction. R2S-EGO instead represents a
behavior-scoped executable query domain within an existing simulator, then
allocates synthetic observations to current support deficits for which
capture-anchored structural conditions are available. Prior work therefore
provides the necessary components---sparse reconstruction, geometry completion,
controlled generation, scene feedback, and active view selection---but does not
jointly define robot-relevant queries and ground the generated observations in
capture-anchored structure.


\section{Method}
\label{sec:method}

\subsection{Problem Setting and Output}
\label{sec:method:overview}

We consider refinement of an existing robot simulation scene. Let
$\mathcal C=\{(I_i,\pi_i)\}_{i=1}^{N}$ denote sparse RGB captures with
calibrated poses. A fixed registration maps the captures and initial asset
$\Ainit$ into the simulator world frame $\mathcal W$, yielding
$\mathcal C^{\mathcal W}$ and $\mathcal A^0$. The simulator $\mathcal S$
supplies the robot embodiment $\mathcal E$, robot dynamics, camera mount
${}^{B}T_C$, and control interface. A behavior scope $\mathcal B$
specifies reachable configurations, admissible motion, and body-relevant
regions. Given $(\mathcal C^{\mathcal W},\mathcal S,\mathcal B,\mathcal A^0)$,
R2S-EGO acquires no additional real observations and returns a refined scene:
the visual asset $\Astar$ together with the current collision surface derived
from the capture-anchored fused geometry
(Sec.~\ref{sec:method:geometry_proxy}). The NKSR complement and collision mesh
are refreshed between rounds; robot dynamics and control interfaces remain fixed.

Useful synthetic observations require two complementary forms of validity.
A robot proxy represents the behavior-scoped executable camera domain and
localizes views that the current visual asset under-renders, while a geometry
proxy supplies capture-anchored structural conditions. A fixed target-view
budget then makes selection necessary: current deficit and geometry availability
prioritize eligible views within the robot-proxy domain. This decomposition
separates the feasibility of a query from the scene support available to answer
it, then couples both before synthesis. We introduce the geometry proxy first
because its availability also enters budgeted selection. The selected target
views
and their rendered structural conditions drive ego-view generation, whose
registered output is assimilated into the visual asset. Throughout, $l$
indexes the refinement round, whether used as a superscript or a subscript; in
particular, $\mathcal A^l$ is the visual asset at the start of round $l$.
Figure~\ref{fig:pipeline} and Algorithm~\ref{alg:r2sego} summarize the pipeline
and its fixed-round execution.

\subsection{Dual-Proxy Ego-View Construction}
\label{sec:method:dual_proxy}

\subsubsection{Capture-Anchored Geometry Proxy}
\label{sec:method:geometry_proxy}

A camera-controlled generator can reproduce a requested motion, but camera
motion alone does not determine the layout, visibility, or occlusion of the
target scene. Sparse captures provide anchors where the scene was observed,
while leaving incomplete support at surfaces revealed from robot views. A
shape prior can supply missing structure, but its prediction begins in an
independent canonical frame. We therefore build the geometry proxy by aligning
prior geometry to capture-derived fragments before using it as a rendering
condition.

Calibrated multi-view tracks are triangulated into fragmentary geometry
$\Mobs=(\mathcal P_{\mathrm{obs}},A_{\mathrm{obs}})$, where the support record
$A_{\mathrm{obs}}$ identifies the captures contributing to each region. A
predefined category vocabulary prompts SAM~3~\cite{Carion2025SAM3}; every
detected instance mask is passed automatically to SAM~3D~\cite{Meta2025SAM3D}
to obtain
$\mathcal P_{\mathrm{prior}}=\{\mathcal P_{\mathrm{prior}}^m\}_m$. The proxy is
then composed as
\begin{equation}
\begin{aligned}
  \Mscaf^{l}=\operatorname{Compose}\!\Bigl(
    &\{T_m^*(\mathcal P_{\mathrm{prior}}^m)\}_m,\\[-1pt]
    &\operatorname{NKSR}\!\bigl(
      \mathcal P_{\mathrm{obs}}^{\neg\mathrm{prior}}
      \cup\mathcal P(\mathcal A^l)^{\neg\mathrm{prior}}\bigr)\Bigr).
\end{aligned}
\label{eq:geometry_proxy}
\end{equation}

$T_m^*(p)=s_m^*R_m^*p+t_m^*$ is the world-frame similarity transform estimated
from overlap between real triangulated and prior points, together with
multi-view consistency between the projected prior silhouette and image masks.
It is solved once from real captures and then held fixed during refinement.
$\operatorname{Compose}$ retains the aligned SAM~3D surface in its object
region; after removing those object points, NKSR reconstructs the remaining
region from real triangulated points and the current backend geometry.
Rendering $\Mscaf^{l}$ at a target pose supplies a frame-aligned structural
condition, including scene layout, visibility, and occlusion. Thus, the
geometry proxy turns an appearance-only generation request into a query tied to
the captured scene, while prior-only surfaces remain structural hypotheses
rather than verified scene geometry.

\subsubsection{Behavior-Scoped Robot Proxy}
\label{sec:method:robot_proxy}

The robot proxy first represents the ego-camera query domain induced by the
robot embodiment and declared behavior scope. In each round, the fixed behavior
controller
$\mu_{\mathcal B}$ is rerun with resampled initial conditions or reference
motions; failed rollouts are discarded and resampled until 12 successful
rollouts are obtained. Forward kinematics and the fixed camera mount yield the
round-specific candidate stream
$\Vcand^l=\{{}^{\mathcal W}T_{C,t}\}_t$. Within this domain, fixed-budget
selection localizes current deficits for which the geometry proxy supplies
usable support. We score each accepted candidate and keep the temporal local
maxima:
\begin{equation}
\begin{aligned}
  {}^{\mathcal W}T_{C,t}&={}^{\mathcal W}T_B(q_t)\,{}^{B}T_C,\\
  s_l(t)&=u_l(t)\,g_l(t),\\
  \Vsel^{l}&=\operatorname{TopK}\!\left(
    \operatorname{NMS}_t(s_l),\Vcand^l,K\right),\quad|\Vsel^{l}|=K.
\end{aligned}
\label{eq:robot_proxy}
\end{equation}

The asset deficit $u_l(t)$ is the fraction of pixels the current visual asset
cannot render at that pose, and the geometry availability $g_l(t)$ is the
fraction the geometry proxy can render there. The values 0.8 in
Table~\ref{tab:reproducibility_settings} are per-pixel alpha/coverage and
depth/splat-validity thresholds used to form these masks, not pose-level gates.
Their product is high only where
visual support is genuinely missing and a usable structural condition exists,
and because the deficit is measured against the current asset, allocation
targets what is missing now rather than global scene coverage. Scoring every
pose leaves long runs of near-identical values, so non-maximum suppression over
$t$ collapses each run into a single keyframe, and $\operatorname{TopK}$ returns
the $K$ highest-scoring surviving poses as the target views. Thus,
$\Vcand^l$ represents the round's behavior-scoped executable query domain,
whereas $\Vsel^{l}$ is only its fixed-budget subset. Each $v\in\Vsel^{l}$ is one
mounted-camera pose the robot actually reaches while performing the declared
behavior, and the fixed controller defines these queries independently of the
downstream learned policy.

\subsubsection{Joint Ego-View Generation}

The robot and geometry proxies contribute complementary motion and structural
inputs, which synthesis must join at a single view index that binds the rendered
structural condition, the reference-relative camera motion, and the generated
frame registration; otherwise the generated frames cannot serve as registered
observations. For each target view $v$, a real capture $I_{i(v)}$ provides the
appearance reference and $\tau_v$ is the camera path along which that view is
generated:
\begin{equation}
\begin{aligned}
  z_v^{l}&=\Renderer(\Mscaf^{l},\tau_v),\\
  X_v^{l}
  &=G_{\theta}\!\left(I_{i(v)},z_v^{l},\bar\tau_v\right),\\
  \Degov^l
  &=\{(X_v^{l,N_f},\hat{\pi}_v^l):v\in\Vsel^{l}\}.
\end{aligned}
\label{eq:dual_proxy_generation}
\end{equation}

Here $N_f$ is the fixed generation sequence length, so $\tau_v$ carries $N_f$
poses, $z_v^{l}$ is the $N_f$-frame structural-condition sequence rendered along
$\tau_v$, and $\bar\tau_v=\{\bar\pi_{v,t}\}_{t=1}^{N_f}$ collects those poses
relative to the reference camera; the rule for $i(v)$, the construction of
$\tau_v$, and the definition of $\bar\pi_{v,t}$ are given in
Sec.~\ref{sec:method:asset_refinement_impl}. Only the terminal frame is retained;
the intermediate frames supply a continuous transition from the reference.
The query $v$ controls generation but is not used as the terminal frame's
training pose. VGGT jointly processes the fixed real captures and generated
frames, estimates each generated pose and intrinsics relative to its real
reference, and composes the relative pose with that reference's fixed world
pose to obtain $\hat{\pi}_v^l$. VGGT estimates for real captures are used only
for this relative composition and never replace their registered poses.

\subsection{Scene Refinement and Implementation}
\label{sec:method:asset_refinement_impl}

Generated frames address missing support only when they become part of the
reusable simulator asset. Let
$\overline{\mathcal D}_{\mathrm{ego}}^l=\bigcup_{j=0}^{l}\Degov^j$ collect
registered generated observations through round $l$. Starting from $\mathcal A^l$,
we refine the visual representation jointly with the registered real captures
and this cumulative observation set. Real captures retain unit weight, while
each generated observation has $w_{\mathrm{syn}}=0.5$ and
$w_{\mathrm{real}}=1$. The captures therefore remain
persistent coordinate and appearance anchors, while the generated observations
extend renderable support at the selected target views. The visual result
retains the input asset interface, while the current fused geometry is installed
as the scene collision surface. Downstream simulation and policy training
therefore consume the refined scene without changing robot dynamics or control
interfaces. Algorithm~\ref{alg:r2sego} gives the complete fixed-round execution
of this output interface.

\begin{algorithm}[!ht]
\caption{R2S-EGO Dual-Proxy Scene Refinement}
\label{alg:r2sego}
\footnotesize
\begin{algorithmic}[1]
  \Require Captures $\mathcal C$; capture registration
           ${}^{\mathcal W}T_{\mathrm{cap}}$; simulator $\mathcal S$; behavior
           scope $\mathcal B$; rounds $L$; target-view budget $K$; initial asset
           $\Ainit$; fixed configuration $\mathcal H$
  \State $(\mathcal C^{\mathcal W},\mathcal A^0)
         \gets\Call{RegisterInputs}
         {\mathcal C,\Ainit,{}^{\mathcal W}T_{\mathrm{cap}}}$
  \State $(\Mobs,\mathcal P_{\mathrm{prior}})
         \gets\Call{ObservedAndPriorGeometry}
         {\mathcal C^{\mathcal W};\mathcal H}$
  \State $\{T_m^*\}_m\gets\Call{AlignPrior}
         {\Mobs,\mathcal P_{\mathrm{prior}};\mathcal H}$
  \Statex \Comment{Prior transforms remain fixed across all rounds}
  \State $\overline{\mathcal D}_{\mathrm{ego}}^{-1}\gets\varnothing$
  \For{$l=0$ to $L-1$}
    \State $\Mscaf^{l}\gets\Call{GeometryProxy}
           {\Mobs,\mathcal P_{\mathrm{prior}},\{T_m^*\}_m,
            \mathcal A^l;\mathcal H}$
    \State $\Call{ReplaceCollisionMesh}{\Mscaf^{l};\mathcal H}$
    \State $\Vcand^l\gets\Call{SuccessfulEgoPoses}
           {\mathcal S,\mu_{\mathcal B},\mathcal B;\mathcal H}$
    \Statex \Comment{Resample failed rollouts until 12 succeed}
    \State $\{i(v)\}_v\gets\Call{SelectReferences}
           {\mathcal C^{\mathcal W},\Vcand^l,\Mobs;\mathcal H}$
    \State $\Vsel^{l}\gets\Call{RobotProxy}
           {\Vcand^l,\Mscaf^{l},\mathcal A^l,K;\mathcal H}$
    \State $\Degov^l\gets\Call{EgoVideoReg}
           {\mathcal C^{\mathcal W},\Mscaf^{l},\Vsel^{l},
            \{i(v)\}_v;\mathcal H}$
    \State $\overline{\mathcal D}_{\mathrm{ego}}^l\gets
           \overline{\mathcal D}_{\mathrm{ego}}^{l-1}\cup\Degov^l$
    \State $\mathcal A^{l+1}\gets\Call{UpdateVisualAsset}
           {\mathcal A^l,\mathcal C^{\mathcal W},
            \overline{\mathcal D}_{\mathrm{ego}}^l;\mathcal H}$
  \EndFor
  \State \Return refined scene: $\Astar\gets\mathcal A^L$ and current collision mesh
\end{algorithmic}
\end{algorithm}

\paragraph*{Registration and geometry}
The fixed transform ${}^{\mathcal W}T_{\mathrm{cap}}$ maps the capture frame to
$\mathcal W$, giving
${}^{\mathcal W}T_{C,i}={}^{\mathcal W}T_{\mathrm{cap}}\pi_i$ and
$\mathcal A^0=\operatorname{RegisterAsset}
(\Ainit,{}^{\mathcal W}T_{\mathrm{cap}})$; ${}^{A}T_B$ maps frame $B$ to frame
$A$. Ground correction rotates the common frame to align its $z$ axis and uses
the first real capture as the new world origin; scale is calibrated separately.
The same transform is applied to captures, assets, robot poses, and collision
geometry. A predefined vocabulary prompts SAM~3, and each detected mask is
passed automatically to SAM~3D \cite{Meta2025SAM3D} using public checkpoints
and default thresholds, without manual instance selection. After prior
alignment, prior-covered object points are excluded from the NKSR input. NKSR
then reconstructs the remaining real points together with current backend
geometry, and its output and the collision mesh are replaced each round.
Prior-completed regions remain structural hypotheses, and we do not
independently evaluate their metric or collision accuracy
(Sec.~\ref{sec:conclusion}).

\paragraph*{Reference-conditioned generation}
For each target view, a fixed rule caches the real reference $i(v)$ with the
greatest capture-anchored overlap, measured as the fraction of proxy points
visible at the target that are also visible in the candidate capture. The path
$\tau_v$ then interpolates from that reference camera to the target pose in
$N_f$ steps so that ViewCrafter
\cite{Yu2024ViewCrafter} receives the continuous transition it requires.
Generator poses are expressed relative to the reference as
$\bar\pi_{v,t}=({}^{\mathcal W}T_{C,i(v)})^{-1}
{}^{\mathcal W}T_{C,v,t}$, and the geometry condition is rendered along the same
path. ViewCrafter receives the cached reference, relative motion, and point
condition; of the returned sequence only the terminal frame, which lies at the
query endpoint, is retained. VGGT then estimates that generated frame's
intrinsics and pose relative to the fixed real reference before it enters the
visual-asset update; the query pose itself is not used as its training pose.

\paragraph*{Backends and controls}
We represent the visual asset with 3DGS. The number of rounds $L$, per-round
target-view budget $K$, generation sequence length $N_f=25$, generation budget,
and reconstruction budget are fixed.
The remaining alignment, rasterization, selection, generation, and
reconstruction settings in $\mathcal H$ are chosen on a disjoint development
split. After each update, support is recomputed over the newly generated
candidate pose stream before the next allocation; the prior transform remains fixed.
The controller rollouts, however, are rerun with newly sampled initial
conditions or reference motions after each geometry/collision update. All
learned components use frozen public checkpoints with no scene-specific fine-tuning.
Comparison arms use the same registered inputs, preprocessing, and simulator
settings. R2S-EGO uses 5k 3DGS steps per round and accumulates at most 36
generated frames over $L=3$ rounds. For every condition that performs
iterative 3DGS optimization, the total reconstruction budget is fixed to 15k
steps, including multi-round and one-shot variants; generated-frame budgets
are additionally matched wherever generation is used. Difix3D+ remains the
fixed official one-step postprocessor and therefore has no iterative
reconstruction budget. R2S-EGO is an offline asset-construction method rather
than a real-time pipeline.

\begin{table}[!t]
  \centering
  \caption{\normalfont\textbf{Key implementation settings.}}
  \label{tab:reproducibility_settings}
  {\footnotesize
  \setlength{\tabcolsep}{3pt}
  \renewcommand{\arraystretch}{1.04}
  \begin{tabular}{@{}p{0.50\columnwidth}p{0.43\columnwidth}@{}}
    \toprule
    Setting & Configuration \\
    \midrule
    Refinement rounds $L$ & 3 \\
    Successful rollouts per round & 12 \\
    Views per round $K$ & 12 \\
    Frames per clip $N_f$ & 25; terminal retained \\
    Synthetic weight $w_{\mathrm{syn}}$ & 0.5; $w_{\mathrm{real}}=1$ \\
    Temporal NMS window & 15 \\
    Asset alpha/coverage pixel threshold & 0.8 \\
    Geometry validity pixel threshold & 0.8 \\
    3DGS optimization & 5k/round; 15k total \\
    Generated-view calibration & VGGT \\
    \bottomrule
  \end{tabular}}
\end{table}

\FloatBarrier


\section{Experiments}
\label{sec:experiments}

We evaluate R2S-EGO through scene refinement and downstream policy use. First,
we compare held-out ego-view appearance against Vanilla 3DGS, GaussGym, and
Difix3D+ on 48 fixed G1-mounted cameras across three Replica scenes. We then
characterize real-image capture-count efficiency on
\texttt{office\_2} with nested input sets, and examine the roles
of SAM~3D geometry grounding, ego-video generation, and repeated scene
refinement through dependency-aware ablations under the common six-view
protocol and 48 frozen target views. Finally, we train otherwise matched
sitting policies in GaussGym and R2S-EGO simulation scenes, report each policy
in its training simulator, and compare them under the same controlled real-G1
deployment protocol.

\begin{figure*}[!t]
  \centering
  \includegraphics[page=6, trim=93 150 96 150, clip,
                   width=\textwidth]{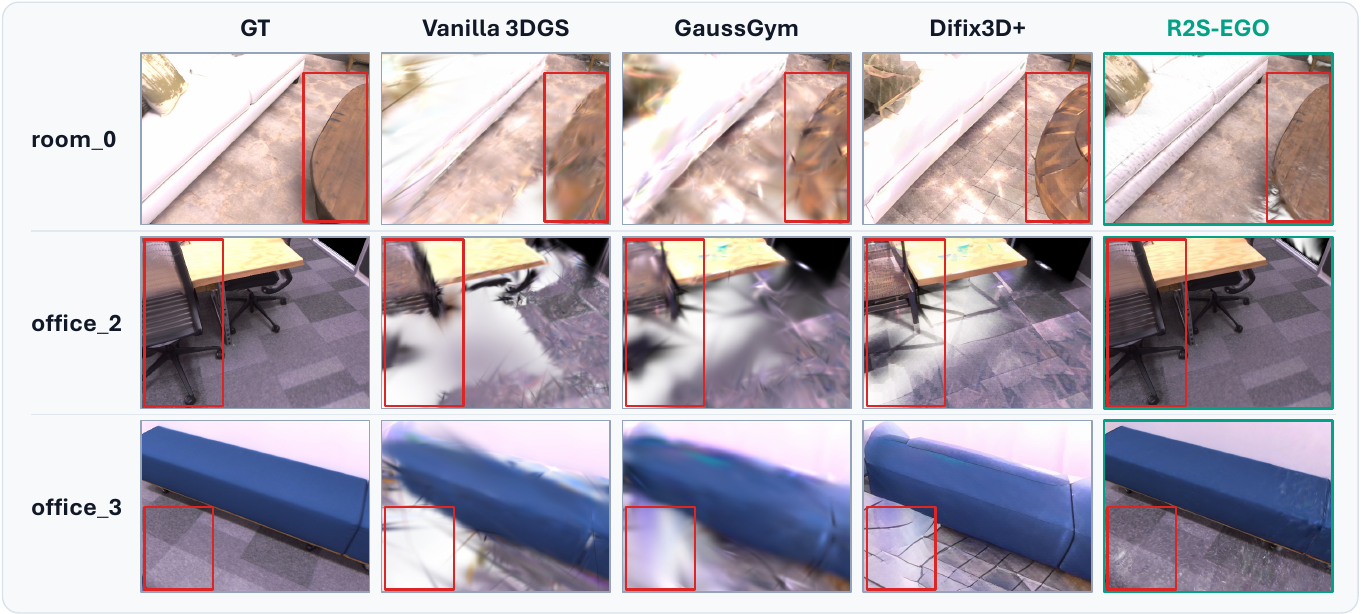}
  \caption{\textbf{Same-camera qualitative comparison on frozen G1 ego views.}
  Each row uses the identical mounted-camera pose for GT, Vanilla 3DGS,
  GaussGym, Difix3D+, and R2S-EGO. The displayed routine-route view is fixed
  within each scene and is not chosen per method. Red boxes mark the same
  image regions in every column: the table rim, the office chair, and the
  sofa boundary, where Vanilla 3DGS and GaussGym degrade into floaters and
  blur and Difix3D+ distorts object structure, while R2S-EGO stays close
  to GT.}
  \label{fig:qual_comparison}
\end{figure*}

\begin{table*}[!t]
  \centering
  \caption{\normalfont\textbf{Quantitative comparison on frozen G1 ego views.}
  Values average 48 views from three Replica scenes. R2S-EGO is the mean over
  five pre-specified pipeline seeds; other methods are single fixed runs.
  Vanilla 3DGS is the formal
  six-view Origin baseline; Difix3D+ is the fixed official one-step
  postprocessor on GaussGym. GenFusion is a component-level visual-update
  comparison rather than an end-to-end R2S system and is evaluated
  under the matched six-view protocol. Best values are bold.}
  \label{tab:frozen_ego_benchmark}
  {\footnotesize
  \setlength{\tabcolsep}{22pt}
  \begin{tabular}{lccc}
    \toprule
    Method & PSNR $\uparrow$ & SSIM $\uparrow$ & LPIPS-Alex $\downarrow$ \\
    \midrule
    Vanilla 3DGS & 13.239 & 0.549 & 0.593 \\
    GenFusion \cite{Wu2025GenFusion} & 14.012 & 0.481 & 0.571 \\
    Difix3D+ & 14.059 & 0.461 & 0.552 \\
    GaussGym & 14.226 & 0.567 & 0.628 \\
    \textbf{R2S-EGO} & \textbf{19.062} & \textbf{0.757} & \textbf{0.273} \\
    \bottomrule
  \end{tabular}}
\end{table*}

\begin{figure}[!htbp]
  \centering
  \includegraphics[width=\linewidth]{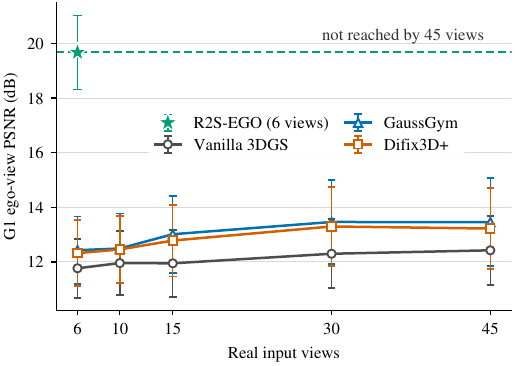}
  \caption{\textbf{Measured real-image capture-count efficiency on G1 ego views.}
  Vanilla 3DGS, GaussGym, and Difix3D+ use nested input-image sets at the five
  displayed budgets; R2S-EGO remains a six-view point. Error bars are SEM over
  16 fixed \texttt{office\_2} target views, not seed uncertainty. No measured
  baseline setting through the 45-view endpoint reaches R2S-EGO@6. GenFusion
  is reserved for the matched six-view component comparison in
  Table~\ref{tab:frozen_ego_benchmark}; because it is not an end-to-end R2S
  baseline, we do not construct a separate capture-count curve for it.}
  \label{fig:dense_view_catchup}
\end{figure}

\subsection{Setup}
\label{sec:exp:setup}

\paragraph*{Benchmark protocol}
We use three Replica scenes: \texttt{office\_2}, \texttt{office\_3}, and
\texttt{room\_0}. Each scene uses six input frames sampled at equal intervals
along a fixed capture trajectory and 16 fixed G1-mounted evaluation cameras: eight
geometry-farthest placements and eight held-out routine-route placements. The
48 targets are distinct from both the inputs and all retained
pseudo-observation target poses. The six inputs, 48 targets, and three
qualitative views were fixed before running any comparison, inspecting method
outputs, selecting hyperparameters, or conducting ablations. Development and
evaluation views are disjoint but come from the same three scenes.
They nevertheless lie within the same declared behavior scope and
robot-mounted camera distribution used by the robot proxy. The benchmark
therefore evaluates generalization to unseen camera poses within the target
behavior distribution, rather than to unseen behaviors, embodiments, or
camera configurations. Exact full-reference metrics require pixel-aligned ground truth at
calibrated held-out robot-camera poses, which is difficult to acquire densely
and repeatably in a physical scene. Replica Ptex rendering provides this
controlled GT, and every GT/render pair uses the same $640\times480$ intrinsics
and camera-to-world matrix. The six real inputs and all evaluation cameras use
Replica ground-truth intrinsics and extrinsics; only generated observations use
VGGT calibration. In physical rooms, where GT calibration is unavailable, VGGT
also estimates the initial real-capture intrinsics and extrinsics. We report
PSNR, SSIM, and LPIPS-Alex over the complete
$640\times480$ images using the original 3DGS metric implementation, with no
mask, crop, exposure/color correction, or image alignment; all pixels are
included and LPIPS uses AlexNet. This benchmark measures controlled rendering
accuracy; the
matched ego-view comparison in Fig.~\ref{fig:real_robot_sitting} and the
real-G1 trials provide complementary real-world evidence.

\paragraph*{Comparison methods}
For the main benchmark and component ablation, all conditions use the same six
registered inputs. All methods use frozen public checkpoints and no
scene-specific fine-tuning. Vanilla 3DGS, GaussGym, Difix3D+, and GenFusion are
reported from one fixed run; full R2S-EGO and every ablation use five
pre-specified, paired pipeline seeds. For seeded results, metrics are averaged
over 48 views within each seed and then across seeds. \emph{Vanilla 3DGS} denotes the formal Origin baseline
reconstructed without generated ego video, obtained by per-scene 3DGS
optimization. The input asset $\mathcal A^0$ that R2S-EGO refines is instead
produced by a feed-forward pointmap-to-Gaussian reconstructor of the
DUSt3R/VGGT family \cite{Wang2024DUSt3R,Wang2025VGGT}, so the \emph{w/o
ego-video generation} ablation row, which reduces to $\mathcal A^0$, is not
numerically identical to Vanilla 3DGS despite using the same six inputs.
\emph{GaussGym} uses its sparse-stable SH3 condition, optimized for the common
15k-step reconstruction budget.
\emph{Difix3D+} is the fixed official one-step postprocessor applied to the
corresponding GaussGym render, not an independent reconstruction.
\emph{GenFusion} is included only in the frozen-view visual benchmark as a
component-level baseline for the visual update: it closes the loop between
video generation and reconstruction, but does not define query allocation from
a fixed robot embodiment and behavior scope or a downstream R2S policy
interface. We evaluate it using the same six registered inputs under a matched
generated-frame and reconstruction budget; it is therefore not
included in the real-G1 policy comparison.
\method{} applies the dual-proxy method in Sec.~\ref{sec:method} using the
fixed-round implementation in Algorithm~\ref{alg:r2sego}.

\paragraph*{Downstream policy and hardware protocol}
We follow GRAIL's egocentric visual sim-to-real setup
\cite{Xie2026GRAIL} on the same SONIC stack, distilling the privileged
scene-aware sitting teacher into a student that maps head RGB and
proprioception to SONIC latent tokens. The GaussGym and R2S-EGO conditions share the teacher, student
architecture, training setup, simulator physics---including the same
final R2S-EGO-derived scene collision mesh and the same set of separately authored
chair collision assets---and SONIC/WBC stack, changing only the training visual
asset; neither uses real-data fine-tuning. Policy training and hardware
evaluation vary the initial positions and orientations of both the robot and
chair and include chair types that differ in backrest, height, shape, and
appearance. At
deployment, the student receives no chair distance, reference ID, or manual
reference initialization. We train five students independently for each
visual-asset condition using five paired policy-training seeds; pair $i$ shares
initialization, teacher data, and training configuration. Seeds, training
steps, and checkpoints are fixed in simulation before any real-robot outcome
is observed. For each seed, we
report 25 in-simulation episodes in its training simulator and eight real-G1
trials with manual resets approximating predefined robot--chair relative-position
and orientation ranges. Methods follow a fixed alternating order,
and the operator is aware of the condition. Success requires no operator
intervention or load-bearing tether engagement and stable chair support for at
least 3 s. We report seed-wise counts
because the independently trained policy, rather than an individual deployment
trial, is the experimental replicate. These hardware trials are intended as a
controlled proof-of-concept validation of the complete R2S-EGO pipeline,
covering robot--chair pose and chair-type variation within the declared sitting
scope. Evaluation of additional behaviors and skills is outside the present
study.

\subsection{Ego-View Quality and View Efficiency}
\label{sec:exp:view_quality}

\paragraph*{Same-view comparison}
Figure~\ref{fig:qual_comparison} compares GT and all four methods at the same
fixed G1-mounted camera in each scene. The red boxes mark identical regions in
every column and concentrate on object boundaries and structure: the
\texttt{room\_0} table rim, the \texttt{office\_2} chair, and the
\texttt{office\_3} sofa contour. There, Vanilla 3DGS and GaussGym collapse
into floaters and blur, and Difix3D+ produces sharp but structurally
distorted objects, whereas R2S-EGO preserves the table rim, chair geometry,
and sofa contour of GT. The
corresponding 48-view aggregate is
reported in Table~\ref{tab:frozen_ego_benchmark}. Among the completed
comparisons, R2S-EGO ranks first on PSNR, SSIM, and LPIPS-Alex, reaching
19.062 dB, 0.757, and 0.273, respectively. The R2S-baseline ordering is
metric-dependent: GaussGym has the strongest PSNR and SSIM, while Difix3D+
has the lowest LPIPS-Alex. GenFusion separately tests whether generic
generation--reconstruction feedback can match the behavior-scoped visual update
without a robot proxy. The
qualitative rows show fixed routine-route views and are not selected separately
for each method.

\paragraph*{Real-image capture-count efficiency}
For the declared \texttt{office\_2} protocol, Vanilla 3DGS,
GaussGym, and Difix3D+ use common nested input-image sets at
$N\in\{6,10,15,30,45\}$ and the same 16 target cameras. Each count keeps the
registered six-view DUSt3R initializer fixed. Both Vanilla 3DGS and GaussGym
use the common 15k-step reconstruction budget, while Difix3D+ postprocesses
the matching GaussGym renders.
R2S-EGO remains a single six-view point. Figure~\ref{fig:dense_view_catchup}
reports PSNR at all five measured budgets, with SEM over the 16 fixed target
views describing target-view variation rather than seed uncertainty. R2S-EGO
obtains 19.674 dB from six real views. At 45 real views, GaussGym, Difix3D+,
and Vanilla 3DGS obtain 13.458, 13.221, and 12.424 dB, respectively. None of
the measured baseline budgets reaches the R2S-EGO six-view level. This result
is specific to the declared \texttt{office\_2} protocol, its 16 frozen target
cameras, and PSNR. We do not
interpolate an unmeasured catch-up count or interpret appearance as geometry
equivalence. GenFusion is evaluated only under the common six-view protocol in
Table~\ref{tab:frozen_ego_benchmark}, where it serves as a component-level test
of generic generation--reconstruction feedback. We therefore neither infer nor
plot an unmeasured GenFusion view-count curve in Fig.~\ref{fig:dense_view_catchup}.

\subsection{R2S-EGO Component Ablation}
\label{sec:exp:component_ablation}

\begin{figure*}[!t]
  \centering
  \includegraphics[page=3, trim=96 203 97 205, clip,
                   width=\textwidth]{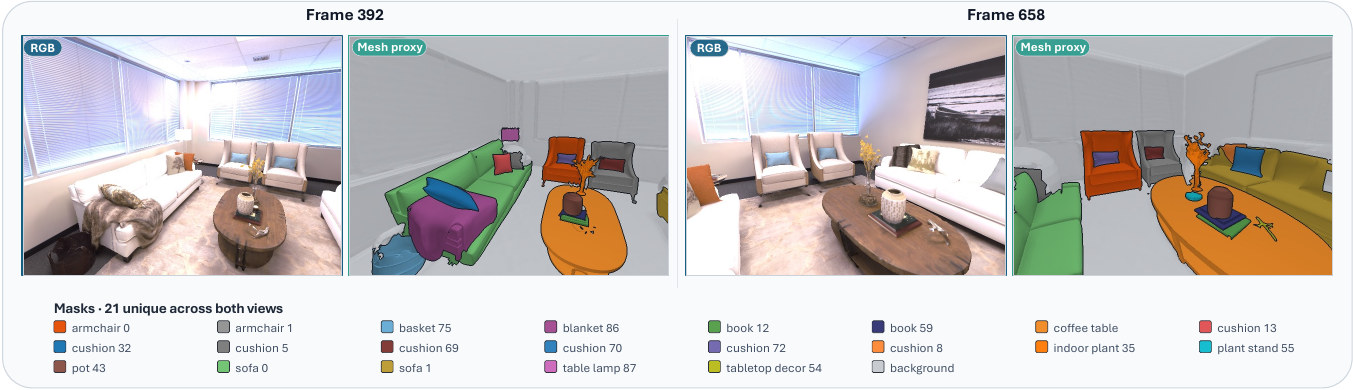}
  \caption{\textbf{Two-view RGB--geometry-proxy correspondence.}
  Each card groups one \texttt{room\_0} view (frames 392 and 658): the RGB
  observation followed by its geometry-proxy rendering. Shared mask colors
  denote object-level associations; the joint legend contains 21 object masks
  plus the background mesh. These annotations are qualitative and are not
  supplied to the generator.}
  \label{fig:mesh_correspondence}
\end{figure*}

\paragraph*{Geometry-proxy visualization}
Figure~\ref{fig:mesh_correspondence} pairs each RGB observation with its
geometry-proxy rendering. Shared colors associate RGB regions with proxy
objects within each pair, and the joint legend lists object IDs across both
views. This visualization is qualitative and supplies no independent
geometry-accuracy metric.

\begin{table}[!htbp]
  \centering
  \caption{\normalfont\textbf{Component ablation of R2S-EGO.}
  Mean over five paired pipeline seeds with six real inputs and 48 frozen G1
  views.}
  \label{tab:r2sego_component_ablation}
  {\footnotesize
  \setlength{\tabcolsep}{3.5pt}
  \renewcommand{\arraystretch}{1.10}
  \begin{tabular}{@{}lccc@{}}
    \toprule
    Variant &
    PSNR $\uparrow$ &
    SSIM $\uparrow$ &
    LPIPS $\downarrow$ \\
    \midrule
    w/o ego video ($\mathcal A^0$)
      & 13.417 & 0.552 & 0.588 \\
    w/o robot-proxy allocation
      & 16.182 & 0.661 & 0.406 \\
    w/o SAM~3D grounding
      & 16.384 & 0.671 & 0.412 \\
    w/o iterative refinement
      & 17.508 & 0.706 & 0.334 \\
    \textbf{Full R2S-EGO}
      & \textbf{19.062} & \textbf{0.757} & \textbf{0.273} \\
    \bottomrule
  \end{tabular}}
\end{table}

\paragraph*{Row-wise ablation protocol}
Table~\ref{tab:r2sego_component_ablation} reports four dependency-aware
variants using the same six inputs and 48 frozen target views.
\emph{w/o SAM~3D geometry grounding} removes the complete object-prior branch;
NKSR alone supplies generation conditioning, collision, and the geometry used
by the next round's rollouts. \emph{w/o ego-video generation} produces no
pseudo-observations and therefore reduces to the feed-forward sparse asset
$\mathcal A^0$. \emph{w/o robot-proxy allocation} retains the fused geometry
condition, generation count, retained-frame count, and optimization budget, but
samples camera-space trajectories without robot feasibility or
deficit/coverage ranking. \emph{w/o iterative scene refinement} selects
36 distinct targets once from the initial proxy, retains 36 terminal frames,
and performs one 15k-step update; it has no intermediate NKSR/collision update,
controller rerollout, or reranking. Its retained-frame count and total 3DGS
optimization budget match full R2S-EGO.

\paragraph*{Component contributions}
Without generated pseudo-observations, the pipeline reduces to $\mathcal A^0$
and loses 5.645 dB PSNR relative to the full method. Replacing robot-proxy
allocation with random generation trajectories lowers PSNR from 19.062 to
16.182 dB and SSIM from 0.757 to 0.661, while LPIPS-Alex increases from 0.273
to 0.406 under matched generation and optimization budgets. Because this
ablation jointly changes candidate-pose construction and ranking, the 2.880 dB
difference supports the coupled robot-proxy allocation procedure as a whole,
rather than isolating the effects of robot feasibility, asset-deficit scoring,
or geometry-availability scoring. Removing SAM~3D grounding gives
16.384 dB, showing that
capture-anchored structural conditioning is complementary to allocation. The
budget-matched one-shot variant reaches 17.508 dB, compared with 19.062 dB for
fixed-round reassessment and refinement. These are aggregate differences over
48 frozen views, not significance or per-scene consistency claims.

\begin{figure*}[!t]
  \centering
  \includegraphics[page=1, trim=97 105 97 105, clip,
                   width=\textwidth]{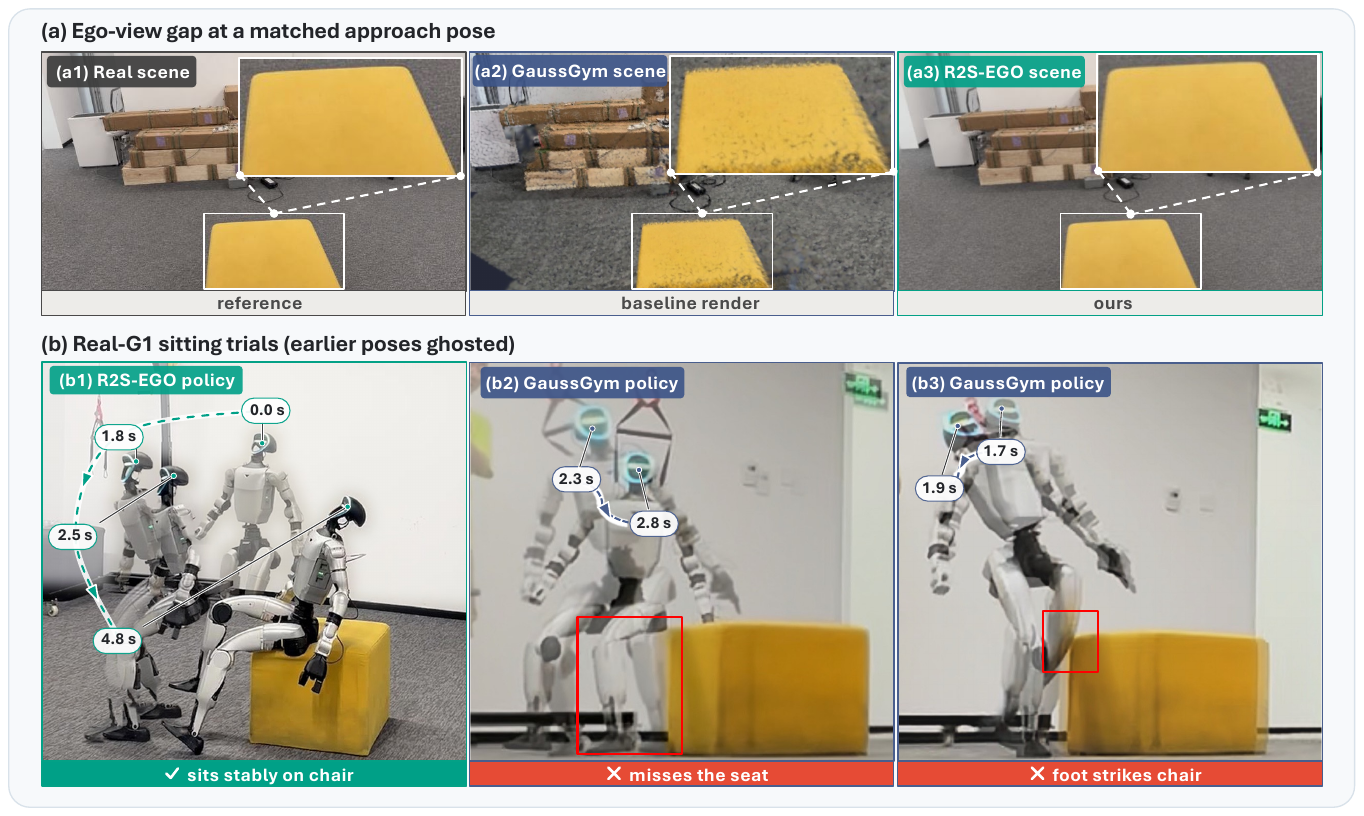}
  \caption{\textbf{Real-G1 sitting validation. The R2S-EGO scene's smaller
  ego-view gap (a) coincides with 33/40 versus 4/40 sitting success across
  five independently trained policies per visual asset (b).}
  (a)~At a matched approach pose, the real head-camera observation (a1)
  versus the same viewpoint rendered from the GaussGym (a2) and R2S-EGO (a3)
  training scenes; a2 is cropped to the real camera's field of view, and the
  boxed chair region is the same in all three and is magnified beside it.
  (b)~Temporally composited real-G1 trials with phase timestamps: an R2S-EGO
  success (b1) and two GaussGym failures whose cause is boxed in red (b2, b3).
  Because (b1) spans the whole approach, its camera sits farther from the chair
  than in (b2) and (b3).}
  \label{fig:real_robot_sitting}
\end{figure*}

\subsection{Real-Robot Sitting Validation}
\label{sec:exp:real_robot}

Table~\ref{tab:realrobot_sitting} reports five independently trained students
per visual-asset condition. Across policy seeds, GaussGym and R2S-EGO obtain
$87.2\%\pm5.2\%$ and $97.6\%\pm2.2\%$ in-simulation success, respectively. On
the real G1, they obtain $10.0\%\pm10.5\%$ and $82.5\%\pm6.8\%$, respectively.
Means and sample standard deviations are computed across independently trained
policies. The hardware difference has the same direction for all five policy
seeds. The pooled counts, 4/40 and 33/40, are descriptive because trials are
nested within policies; the paired real-G1 seed counts are
$(0,0,1,2,1)$ versus $(7,7,7,6,6)$. Similar in-simulation performance and consistently
separated hardware results support an association between the refined ego-view
asset and transfer under this protocol.
Figure~\ref{fig:real_robot_sitting}(a) provides a same-pose qualitative
comparison between the real head-camera observation and the two training-asset
renderings. Figure~\ref{fig:real_robot_sitting}(b) shows a time-annotated
R2S-EGO success and two representative GaussGym failures whose annotated
causes---a premature sit-down that misses the seat and a foot--chair
collision during approach---are illustrative examples rather than a causal
diagnosis of the success-rate gap. Each policy first runs with a slack overhead
safety tether. The tether supplies no support during nominal motion and becomes
load-bearing only after loss of balance; tether engagement or operator
intervention is counted as failure. A policy is detached only after a successful
tethered sit. The illustrated GaussGym policy did not pass this gate, so its
failure trials remain tethered, whereas the R2S-EGO success in
Fig.~\ref{fig:real_robot_sitting}(b1) executes untethered.
Across the evaluated robot--chair configurations and chair types, R2S-EGO
achieves higher hardware success for each policy seed.

\begin{table}[!h]
  \centering
  \caption{\normalfont\textbf{Success rates in simulation and real-world deployment.}
  Mean $\pm$ sample standard deviation across five independently trained
  policies; 25 simulation episodes and 8 real-G1 trials per policy,
  spanning varied robot--chair poses and chair types.}
  \label{tab:realrobot_sitting}
  {\footnotesize
  \setlength{\tabcolsep}{3.5pt}
  \renewcommand{\arraystretch}{1.10}
  \begin{tabular}{@{}lcc@{}}
    \toprule
    Training scene & Sim (\%, mean $\pm$ std) &
    Real (\%, mean $\pm$ std) \\
    \midrule
    GaussGym & 87.2 $\pm$ 5.2 & 10.0 $\pm$ 10.5 \\
    \textbf{R2S-EGO} & \textbf{97.6 $\pm$ 2.2} &
    \textbf{82.5 $\pm$ 6.8} \\
    \bottomrule
  \end{tabular}}
\end{table}


\section{Conclusion}
\label{sec:conclusion}

We presented R2S-EGO, a dual-proxy method for R2S scene refinement along
behavior-scoped robot views in a sparse-capture scene. Rollout-derived robot
poses define the behavior-scoped executable query domain, and current-deficit
scoring identifies which ego views to generate; capture-anchored geometry
supplies structural conditioning and the refreshed collision surface. The
resulting registered generated observations are assimilated at lower weight
into the visual asset. Across 48
frozen G1 ego views in three Replica scenes, six-view R2S-EGO reaches
19.062 dB PSNR versus 14.226 dB for the strongest reported R2S baseline. Across
five independently trained policies per condition, GaussGym and R2S-EGO
succeed in 4/40 and 33/40 controlled real-G1 sitting trials, respectively, with
the same ordering for every seed. These results support targeting
sparse-capture scene refinement to behavior-scoped robot views to improve
robot-view rendering and downstream policy performance in the evaluated
settings.

Three boundaries remain. Prior-completed geometry is a capture-anchored
structural hypothesis rather than scene truth; although the fused geometry
also instantiates the scene collision surface used downstream, we do not
independently quantify its metric or collision accuracy, nor isolate its
contribution from visual-asset refinement. The robot proxy covers a
declared behavior scope rather than learned-policy occupancy, and repeated
pseudo-observation updates may propagate errors in generated appearance or
prior-completed structure. As a proof-of-concept complement to the main visual
benchmark, the hardware evaluation covers variation in robot and chair pose and
in chair backrest, height, shape, and appearance across five independently
trained policies per visual asset. Evaluation of additional behaviors and
skills, including arm-based manipulation and fine-grained hand--object
interaction, remains future work.


\FloatBarrier
\bibliographystyle{IEEEtran}
\bibliography{references}

\end{document}